# PMKLC: Parallel Multi-Knowledge Learning-based Lossless Compression for Large-Scale Genomics Database

Hui Sun
sunh@nbjl.nankai.edu.cn
College of C.S., Nankai Uiversity &
College of CCSD, Nanyang
Technological University
Tianjin, China & Nanyang, Singapore

Yanfeng Ding
dingyf@nbjl.nankai.edu.cn
College of C.S., Nankai-Baidu Joint
Lab, TMCC, SysNet, DISSec, GTIISC,
Nankai Uiversity
Tianjin, China

Liping Yi
yiliping@nbjl.nankai.edu.cn
College of C.S., Nankai Uiversity &
College of CCSD, Nanyang
Technological University
Tianjin, China & Nanyang, Singapore

Huidong Ma
mahd@nbjl.nankai.edu.cn
College of C.S., Nankai-Baidu Joint
Lab, TMCC, SysNet, DISSec, GTIISC,
Nankai Uiversity
Tianjin, China

Gang Wang*
wgzwp@nbjl.nankai.edu.cn
College of C.S., Nankai-Baidu Joint
Lab, TMCC, SysNet, DISSec, GTIISC,
Nankai Uiversity
Tianjin, China

Xiaoguang Liu*
liuxg@nbjl.nankai.edu.cn
College of C.S., Nankai-Baidu Joint
Lab, TMCC, SysNet, DISSec, GTIISC,
Nankai Uiversity
Tianjin, China

Cheng Zhong
chzhong@gxu.edu.cn
School of Computer, Electronics and
Information, Guangxi University,
Nanning, China

Wentong Cai
ASWTCAI@ntu.edu.sg
College of CCSD, Nanyang
Technological University
Nanyang, Singapore

## Abstract

Learning-based lossless compressors play a crucial role in large-scale genomic database backup, storage, transmission, and management. However, their 1) inadequate compression ratio, 2) low compression & decompression throughput, and 3) poor compression robustness limit their widespread adoption and application in both industry and academia. To solve those challenges, we propose a novel <u>P</u>arallel <u>M</u>ulti-<u>K</u>nowledge <u>L</u>earning-based <u>C</u>ompressor (PMKLC) with four crucial designs: 1) We propose an automated multi-knowledge learning-based compression framework as compressors' backbone to enhance compression ratio and robustness; 2) we design a GPU-accelerated ($s,k$)-mer encoder to optimize compression throughput and computing resource usage; 3) we introduce data block partitioning and Step-wise Model Passing (SMP) mechanisms for parallel acceleration; 4) We design two compression modes PMKLC-S and PMKLC-M to meet the complex application scenarios, where the former runs on a resource-constrained single GPU and the latter is multi-GPU accelerated. We benchmark PMKLC-S/M and 14 baselines (7 traditional and 7 leaning-based) on 15 real-world datasets with different species and data sizes. Compared to baselines on the testing datasets, PMKLC-S/M achieve the average compression ratio improvement up to 73.609% and 73.480%, the average throughput improvement up to 3.036× and 10.710×, respectively. Besides, PMKLC-S/M also achieve the best robustness and competitive memory cost, indicating its greater stability against datasets with different probability distribution perturbations, and its strong ability to run on memory-constrained devices. Overall, PMKLC is a balanced compression solution that optimizes compression ratio, throughput, robustness, and resource consumption. PMKLC and linkages of datasets are available at https://github.com/dingyanfeng/PMKLC.

## 1 Introduction

The creation of Large-Scale Genomic Databases (LSGD) poses challenges for data storing, sharing, and managing. For example, as of January 2025, the China National GeneBank DataBase (CNGB) has backed up 17,267 TB of big genomics data [1]. The development of dedicated Learning-based Lossless Compressors (LLCs) for LSGD

---

is the main approach to alleviating these pressures [2–5]. Existing LLCs belong to statistical-based compression methods that achieve lossless compression by constructing a neural-network (NN)-based model and combining it with an entropy encoder [6]. Compared to traditional dictionary-based methods [2, 7, 8], existing LLCs for LSGD have advantages in reducing the size of genomic data, but they still face some drawbacks and challenges.

**Challenge1: Inadequate Compression Ratio.** This is caused by three reasons. 1) The cold-start problem [4, 9], where the training of the models of LLCs is insufficient in the initial batches, leads to ineffective compression by the compressor in the initial rounds. 2) Insufficient learning of the dataset to be compressed by the compressor, which means the sources of knowledge are not adequate. 3) Simple model design, which cannot fully exploit and utilize the redundant information in genomics data.

**Challenge2: Low Throughput.** Existing LLCs have low compression & decompression throughput, making them unsuitable for medium-to-long-term backup compression scenarios when compressing and managing large-scale genomics data. For example, in a HuMa dataset [10], DeepDNA [3], DNA-BiLSTM [11], and Gene-Former [12] achieve throughput by 5.409, 4.754 and 3.190 KB/s, while non-learning-based MFCompress [13] and GenoZip [14] reaches 10.134 and 6.699 MB/s, respectively. This is mainly due to two reasons: 1) the high cost of learning and inference in deep learning models, and 2) insufficient parallel mechanisms.

**Challenge3: Poor Compression Robustness.** Most LLCs use fixed model architecture designs, resulting in poor compression robustness. For example, the traditional static LSGD compressors need to save the model as part of the compressed file. When the data size is small, the saved model becomes a burden on the compression process, leading to poor robustness and generalization.

To address those issues, this paper proposes a novel Parallel Multi-Knowledge Learning-based Compressor (PMKLC) for large-scale genomic databases lossless compression, aiming to achieve an overall balance of compression ratio, throughput, robustness, and memory usage. Our contributions are summarized as follows:

**1)** We propose an Automated Multi-Knowledge Learning-based Compression Framework (AMKLCF) as the compressors' backbone, which consists of three key NN-based models: a) a Static Public Model (SPuM) for learning multi-genomic data knowledge, b) a Static Private Model (SPrM) for learning the global knowledge of the to-be-compressed dataset, and c) a Dynamic Model (DM) for self-learning as well as integrating multi-source knowledge. Through automated model selection, AMKLCF enhances the compression ratio and robustness.

**2)** We design a GPU-accelerated ($s,k$)-mer Encoder (GskE) that effectively extracts redundant information from the genomics data and reduces the data size, further enhancing throughput and compression ratio as well as alleviating memory usage. We also use data block partitioning and a Step-wise Model Passing (SMP) mechanism, employing multi-GPU parallelism to accelerate the process of compression and decompression.

**3)** We design two compression modes: a) PMKLC-S, which is accelerated by a single GPU, and b) PMKLC-M, which is accelerated by multiple GPUs for higher throughput. Extensive experiments show that both PMKLC-S and PMKLC-M achieve the best overall compression ratio, throughput, robustness, and memory cost.

The rest of the paper is organized as follows: section 2 gives the background of the learning-based compressor; section 3 provides related work; section 4 shows the designs of PMKLC; section 5 gives experimental results and analysis; and section 6 is brief conclusions.

## 2 Preliminary

LLCs for genomics data belong to statistical-based methods [15–19], including two components: NN-based modeling and entropy coding [2, 20]. Let $X = \{x_0, x_1, .., x_{n-1}\}$ denotes the to-be-compressed genomics sequence with length $n$. For the compression process, a typical learning-based compressor includes the following steps:

(1) Initializing NN-based model $\mathcal{M}$ and entropy encoder $\mathcal{E}$.
(2) Dividing $X$ into **historical symbols** $\{x_{i-t}, ..., x_{i-1}\}$ and the to-be-compressed **target symbol** $x_i$ based on the **context length** $t$, where $i = \{t, ..., n-1\}$.
(3) For each $\{x_{i-t}, ..., x_{i-1}\}$, using $\mathcal{M}$ to calculate the **probability distribution** $p(x_i|x_{i-t}, ..., x_{i-1})$ of target symbol $x_i$, and pass it to $\mathcal{E}$ to obtain the compressed binary stream $c_i$.
(4) Calculating the cross entropy loss using $p(x_i|x_{i-t}, ..., x_{i-1})$ and $x_i$, and updating the parameters of model $\mathcal{M}$ using the back-propagation algorithm to improve the model's predictive ability, where $i = \{t, ..., n-1\}$.

In step (3), for the initial $t$ symbols, the LLC calculates their uniform distribution [20–22] and passes it to the encoder $\mathcal{E}$. Additionally, based on whether model $\mathcal{M}$ is updated during the compression process in step (4), existing LLCs can be classified into Static, Dynamic, and Semi-adaptive, as shown in Fig. 1. This will be detailed in the next section.

For the decompression process, the LLC first restores the first $c$ target symbols through the decoder $\mathcal{E}$ and uniform distribution. Then, the sequence is restored one by one through model $\mathcal{M}$, where $i = \{t, ..., n-1\}$. Since the decompressor employs the same model $\mathcal{M}$ as the compressor, this ensures that no information is lost during the decompression process, thus allowing the LLC to restore the original data losslessly.

## 3 Related Work

Fig. 1 presents an overall schematic of three LLCs compression frameworks. Since the general-purpose LLCs for multi-source data [20, 21, 23] are also applicable to Genomics Data (GD) compression, we will also provide a brief introduction to it.

### 3.1 Static

As shown in Fig. 1(a), a Static LLC includes two stages, it first pre-trains a static model $\mathcal{M}_{static}$, and then uses it to compress GD. Since the model is pre-trained, it needs to be saved as part of the compressed file. The main differences between current Static LLCs lie in the models' architecture. DeepDNA [3] and DNA-BiLSTM [11] are the earliest static LLCs, using LSTM [24] and BiLSTM integrated with attention mechanisms [25] as compression models, respectively. Following, CompressBERT [26], GenCoder [27], Gene-Former [12], LEC-Codec [28] are proposed, using BERT [29] language model, convolutional auto-encoder, Transformer [30] with multi-level-grouping, and Group of Bases (GoB) compression with LSTM, respectively. Additionally, some general-purpose compressors for multi-source data have also been used to compress LSGD, including Large Language Model (LLM)-based LLMzip [31], BiGRU-based DeepZip [23], and LSTM-based Lstm-compress [32], et al.

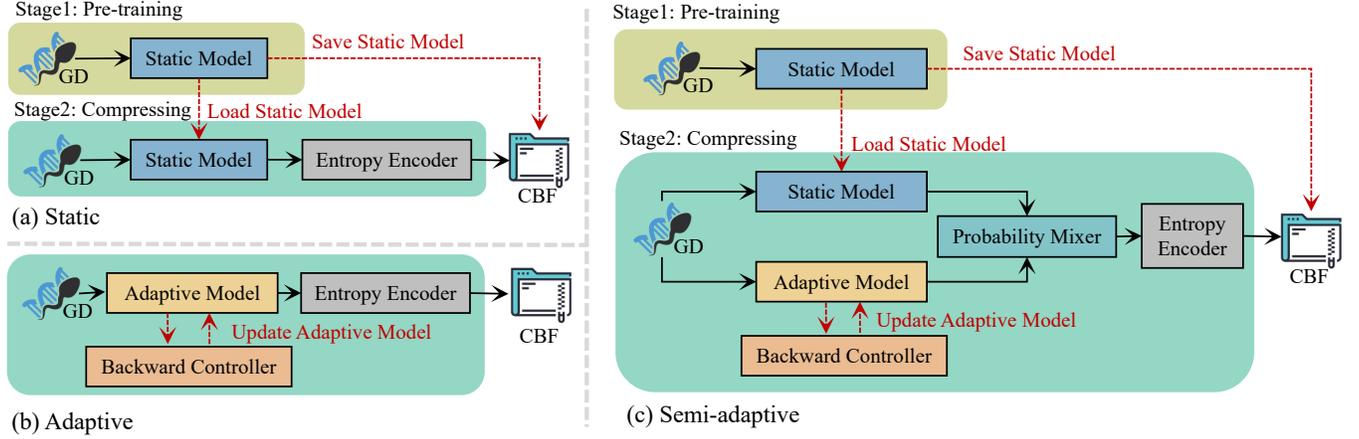

Figure 1: The compression diagrams for Static (a), Adaptive (b), and Semi-adaptive (c) compression frameworks. Here, GD and CBF denote the Genomics data and Compressed Binary File, respectively

Table 1: Comparison of Static, Adaptive, and Semi-adaptive frameworks. More "⋆" indicate a stronger advantage

| Framework | Compression Ratio | Throughput | Compression Robustness |
|---|---|---|---|
| Static | ⋆⋆⋆ | ⋆ | ⋆ |
| Adaptive | ⋆ | ⋆⋆⋆ | ⋆⋆⋆ |
| Semi-adaptive | ⋆⋆ | ⋆⋆ | ⋆⋆ |

### 3.2 Adptive

Adaptive LLCs do not require any pre-training. As shown in Fig. 1(b), they dynamically update the adaptive model $M_{adaptive}$ during the data compression process. AGDLC [33] is the first Adaptive LLC for genomics data, combining the advantages of multiple (s,k)-mer encoding [34] and XLSTM [35]. DeepGeCo (PLUS Mode) [4] is a flexible GD compressor that uses Transformer [20, 30] as the backbone and introduces a BiGRU-based module [22, 23] to address the cold-start problem. Besides, general-purpose compressors like XLSTM-based MSDLC, Transformer-based TRACE [20], MLP-based OREA [36] and PAC [21] also perform well in GD compression.

### 3.3 Semi-Adaptive

Semi-adaptive LLCs have not been well explored in the field of GD compression, with the known DZip (Supporter Mode) [22] being its only proponent. Taking Fig. 1(c) as an example, Semi-adaptive LLC first pre-trains a static model $M_{static}$ on GD, then uses a preset adaptive model $M_{adaptive}$ and $M_{static}$ together to perform probability prediction through a Probability Mixer (PM). Meanwhile, since $M_{static}$ is pre-trained, it does not need to be updated, while $M_{adaptive}$ needs to update its parameters through a Backward Controller (BC). In Dzip, $M_{static}$ and $M_{adaptive}$ are implemented based on BiGRU and MLP, respectively.

### 3.4 Brief Summary

We summarize the strengths and weaknesses of the three frameworks (assuming the same model scale), as shown in Table. 1.

For Static LLCs, the main strength is compression ratio, especially for large-scale data. However, when the dataset is small, the model becomes a burden on compression, leading to poorer compression robustness. Additionally, since static LLCs require pre-training, they do not have a time advantage. Adaptive LLC does not require pre-training or saving model parameters, so it has better compression throughput and robustness, but at the cost of compression ratio degradation. The design principle of Semi-adaptive is to combine the advantages of Static and Adaptive methods, that is, to use a smaller $M_{static}$ for pre-training to save time, and further improve the compression ratio through $M_{adaptive}$. Therefore, its performance is overall balanced. However, it still inherits the drawbacks of both Static and Adaptive methods, namely slower throughput and robustness compared to Adaptive, and poorer compression ratio compared to Static.

In this paper, we propose a novel Automated Multi-Knowledge Learning-based Compression Framework (AMKLCF), which differs from the aforementioned frameworks. On top of AMLCF, we introduce a GPU-based (s,k)-mer encoder and multi-GPU parallelism to further optimize compression ratio, compression & decompression throughput, compression robustness, and memory consumption.

## 4 PMKLC Design

Fig. 2 shows the automated compression pipeline of PMKLC using 2 GPUs, it consist of three crucial stages:

**a) MSGD Pre-training.** It aims to pre-train a multi-knowledge-driven Static Public Model (SPuM) on MSGD (Multi-Source Genomics Data). This motivation stems from the similarity in genomics data among homologous species. For example, the genetic similarity between humans and gorillas is as high as 99.5% [37, 38]. Thus, a model that includes the probability of gorilla genes would help improve the compression of human genomics data [4].

**b) Static Pre-training.** This stage is designed similarly to the pre-training phase of the Semi-adaptive compression framework

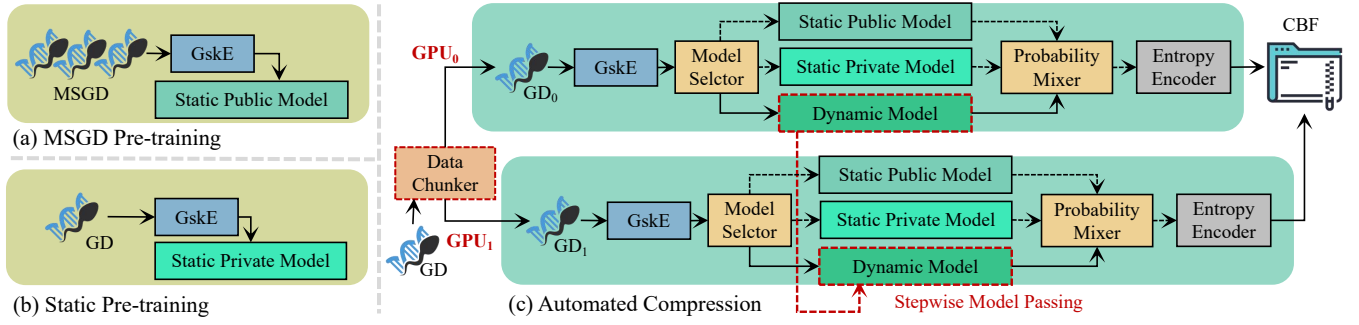

Figure 2: Compression pipeline of the proposed PMKLC using 2 GPUs. Here, GD, MSGD, CBF, and GskE denote Genomics Data, Multi-Source Genomics Data, Compressed Binary File, GPU-based (s,k)-mer Encoder, respectively

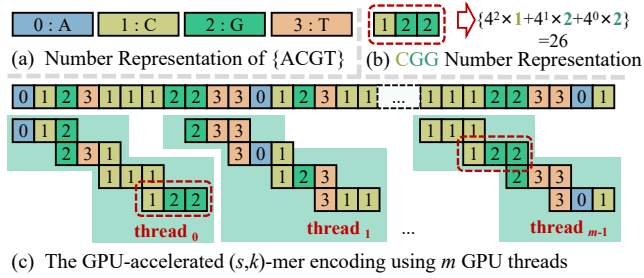

Figure 3: The GPU-based (s,k)-mer Encoder (GskE)

as Fig. 1(c) shows. It is used to learn the global features of the to-be-compressed genomics data and generate a Static Private Model (SPrM) for further enhancing the compression ratio.

c) **Automated Compression.** It is based on a self-learning Dynamic Model (DM) and combined with the multi-knowledge of SPuM and SPrM for compressing. A group of Model Selector (MS) and Probability Mixer (PM) are designed to select knowledge and mix probabilities from SPuM, SPrM, and PM. We call the combination of these models and components as the Automated Multi-Knowledge Learning-based Compression Framework (AMKLCF). On top of it, a GPU-based (s,k)-mer encoder (GskE) is designed to improve compression ratio and throughput. A group of Data Chunker (DC) and Stepwise Model Passing (SMP) mechanism are introduced to increase parallelism using multiple GPUs. Besides, the Entropy Encoder (EE) is used to generate a Compressed Binary File (CBF) from the PM.

In the following subsections, we will detail the key components GskE, AMKLCF, DC, and SMP mechanism as shown in Fig 2(c).

### 4.1 GPU-based (s,k)-mer Encoder

GskE reduces the original data to a smaller scale to improve throughput, as shown in Fig. 3. Additionally, since GD is composed only of {ACGT} and has a high degree of redundancy, GskE also leverages this redundancy to enhance the compression ratio. GskE's design is similar to dictionary-based encoding [7, 39], with $k$ representing the window size. However, by introducing the step size parameter $s$, GskE offers greater flexibility. Appropriate parameter settings allow the compressor to better balance compression ratio and throughput.

As shown in Fig. 3(a), GskE first translates GD to number representation, that is {A:0, C:1, G:2, T:3}. Following, it calculates the encoded number representation. As shown in Fig. 3(b), the substring "CGG" is encoded as $4^2 \times 1 + 4^1 \times 2 + 4^0 \times 2 = 26$. Finally, in Fig. 3(c), the multiple GPU threads are used for parallel calculation with the widows size $k = 3$ and step size $s = 2$.

### 4.2 Automated Multi-Knowledge Learning-based Compression Framework

Fig. 4 shows the AMKLCF framework of the PMKLC compressor. It takes the (s,k)-mer encoded vector $E$ as input, and generates a mixed probability vector $P$ as output. Here, we introduce some crucial designs of the proposed AMKLCF.

*4.2.1 **Model Selector (MS)**.* AMKLCF takes DM as the basic backbone and combines SPuM and SPrM to enhance the compression ratio. While this enriches the knowledge sources to some extent, it leads to excessive resource consumption. Additionally, we found that the combination of the three models is sensitive to data scale (see Ablation Study). Therefore, we design MS to enable SPuM ($\leq$ 500MB) or SPrM ($>$ 500MB) automatically based on the size of the dataset to balance resource consumption and throughput. The MS takes $E$ as input and generates a binary vector $S$ as output. For example, $S = \{1, 0, 1\}$ means automatically enabling SPuM and DM.

*4.2.2 **Static Public Model (SPuM)**.* It is a well-trained static model on multi-source GD for solving cold-start problems on small-size data, like DeepGeCo [4]. As a hard-coded component of PMKLC, SPuM is frozen during the compression process and does not require parameter updates. Let $E = \{e_0, ..., e_{l-1}\}$ denotes the $(s, k)$-mer encoded vector with length $l$ and where $e \in \mathbb{R}^{1 \times 4^k}$, $t$ denotes the context length. For each historical symbol in $E_i = \{e_{i-t}, ..., e_{i-1}\}$, $i = \{t, ..., l-1\}$, DM extracts GD redundancy via a 16-dimension Embedding layer, 2-BiGRU layers, and a Linear and Dense layer with a dimension of 128. After that, it generates a learned knowledge vector $L_i^u$ as output, where $L_i^u \in \mathbb{R}^{1 \times 4^k}$.

*4.2.3 **Static Private Model (SPrM)**.* SPrM shares a similar model structure and parameters as SPuM, with the only difference being a single-layer BiGRU and take $L_i^r$ as output. This is for the following three reasons: 1) Shared parameters facilitate the implementation of the automated model selection framework. 2) It reduces time consumption, as SPrM requires pre-training. 3) It reduces the model size's proportion in the compressed file, as SPrM needs to be preserved to losslessly restore the original data.

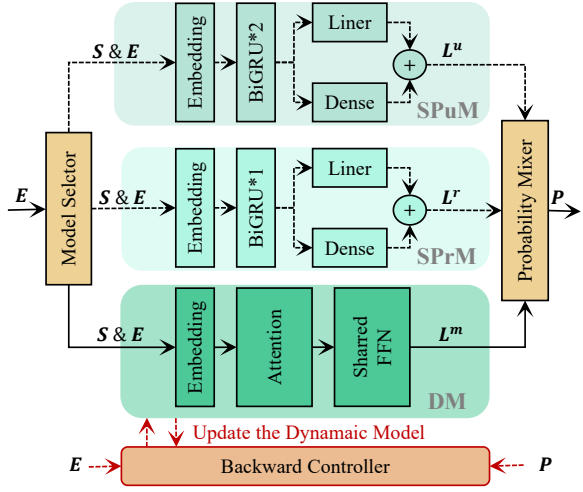

Figure 4: The Automated Multi-Knowledge Learning-based Compression Framework (AMKLCF). $E$, $S$, $L^{\{u,r,m\}}$, and $P$ denote the (s,k)-mer encoded data, selected models, learned knowledge from each model, and the final probability

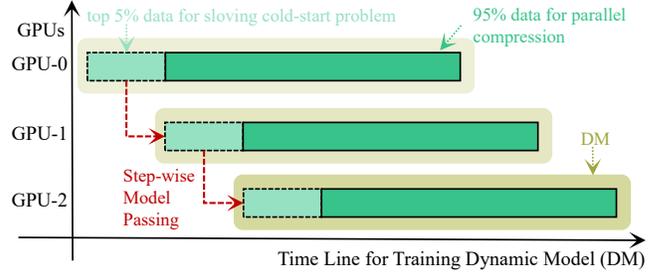

Figure 5: Parallelization using Step-wise Model Passing (PSM) strategy and multiple GPUs

*4.2.4 Dynamic Model (DM).* This is the largest of the three models in parameters. During the compression process, it dynamically updates the model through a Backward Controller (BC), so it does not need to be saved as part of the compressed file. It consists of a 64-dimension Embedding layer, an 8-head Attention layer with a hidden dimension of 256, and a 4096-dimension shared FFN (Feed-Forward Network) layer. The DM generates a learned vector $L_i^m$ as output, which is fed into the Probability Mixer, where $L_i^m \in \mathbb{R}^{1 \times 4^k}$.

*4.2.5 Probability Mixer (PM).* It mixes the learned knowledge and produce the final probability $P_i = P(e_i|e_{i-t},...,e_{i-1})$ via a Softmax function, here $P_i \in \mathbb{R}^{1 \times 4^k}$ and $i = \{t,...,l-1\}$. Probability mixing is an automated learning process that combines the knowledge of multiple models, calculated as follows:

$$P_i^s = Softmax\left(\alpha \times (S_0 \times L_i^u + S_1 \times L_i^r) + (1-\alpha) \times S_2 \times L_i^m\right), \quad (1)$$

where, $\alpha$ is a self-learning weight parameter of the framework, and $S_0$, $S_1$ and $S_2$ are model flags generated by the Model Selector.

*4.2.6 Backward Controller (BC).* This module updates the DM via a Cross-Entropy [4, 11, 40] function. For each one-hot encoded ground truth label $E_i \in \mathbb{R}^{1 \times 4^k}$ of $e_i$ and predicted probability vector $P_i \in \mathbb{R}^{1 \times 4^k}$ of $\{e_i|e_{i-t},..,e_{i-1}\}$, the loss is calculated as:

$$\mathcal{L}(E,P) = \sum_{i=t}^{l-1} CE(E_i, P_i) = \sum_{i=t}^{l-1} \sum_{j=0}^{4^k-1} (E_{ij} log \frac{1}{P_{ij}}). \quad (2)$$

PMKLC focuses on the overall design of genomic data compression frameworks. We explored the design of multiple model architectures, with the BiGRU-based SPrM and SPuM, and the Transformer-based DM being the most effective [4, 20, 22, 23].

### 4.3 Parallel Acceleration

To further enhance throughput, we employ data block partitioning strategy to evenly partition the raw data, which is then accelerated using multiple GPUs. However, multi-GPU acceleration exacerbates the cold-start problem [4, 9], as each GPU needs to maintain its own unique Dynamic Model. To address this issue, we introduce a Step-wise Model Passing (SMP) strategy [9]. Taking Fig. 5 as an example, GPU-0 trains the model DM-0 from randomly initialized parameters. After training on 5% of the data, DM-0 is passed to GPU-1 as DM-1. This avoids initializing DM-1 from random parameters, solving the cold-start problem for GPU-1 in the initial batch compression and further enhancing the compression ratio.

Our PMKLC is the first LSGD compressor accelerated by multiple GPUs. To meet complex applications in GPU resource-constrained scenarios, we also design two compression modes: 1) PMKLC-S, which operates on a single GPU, and 2) PMKLC-M, which uses data block partitioning and SMP strategies for multi-GPU acceleration.

## 5 Result

All experiments were conducted on a Ubuntu server (20.04.6 LTS) equipped with 4 × Intel Xeon 4310 CPUs (2.10 GHz), 4 × NVIDIA GeForce RTX 4090 GPUs (24 GB), and 512 GB of DDR4 RAM.

### 5.1 Setup

*5.1.1 Datasets.* As shown in Table 2, we used 15 open-source real-world datasets for experimental evaluation, which include multiple species and various data scales [1, 2, 4, 10, 37, 38].

*5.1.2 Baselines.* We compared PMKLC with 14 baselines, including 7 learning-based and 7 traditional non-learning-based methods. Table 3 shows the published or source-code updated year of baselines, programming language, and major technologies.

*5.1.3 Metrics.* We compared PMKLC and baselines on multiple public metrics, consisting of compression ratio [49–51]:

$$CR = \frac{Size_{compressed}}{Size_{source}} \times 8\ bits/base, \quad (3)$$

Table 2: Details of used datasets

| Name | Size (Bytes) | Description |
|---|---|---|
| **SPuM Training and Parameters Tuning:** | | |
| DNACorpus | 685,597,124 | compression corpus with multi-species GD |
| OrSa | 43,262,523 | the chromosome-1 of oryza sativa japonica |
| DiCo | 554,197,554 | a collection of unknown genomics data |
| EnIn | 26,403,087 | one set of genome from entamoeba invadens |
| ScPo | 10,652,155 | the chromosome-3 of a beer yeast |
| DaRe | 62,565,020 | the chromosome-3 of the danio rerio |
| **Compression Testing:** | | |
| PlFa | 8,986,712 | one set of genome from plasmodium falciparum |
| WaMe | 9,144,432 | a collection of unknown genomics data |
| DrMe | 32,181,429 | the chromosome-2 of the drosophila miranda |
| GaGa | 148,532,294 | the chromosome-2 of the gallus gallus |
| SnSt | 6,254,100 | one collection of STRs and SNPs of human |
| MoGu | 171,080,468 | the genomics data of mouse gut |
| ArTh | 19,695,740 | part of data from arabidopsis thaliana |
| AcSc | 714,975,658 | a set of acanthopagrus schlegelii genomics data |
| TaGu | 1,052,636,474 | a set of taeniopygia guttata genomics data |

Table 3: Details of compared baselines

| Name | Year | Language | Technologies |
|---|---|---|---|
| **Learning-based Compressors:** | | | |
| MSDLC [41] | 2025 | Python | XLSTM [35], Expansion Mapping |
| AGDLC [33] | 2025 | Python | XLSTM, Multi-$(s,k)$-mer Encoding |
| JARVIS3 [42] | 2024 | C/C++ | Finite-context and Repeat Models |
| TRACE [20] | 2022 | Python | Attention, Transformer |
| DNA-BiLSTM [11] | 2018 | Python | Attention, BiLSTM, CNN |
| DeepZip [23] | 2018 | Python | LSTM, BiGRU |
| Lstm-compress [32] | 2017 | C/C++ | LSTM, MLP |
| **None-learning-based Compressors:** | | | |
| LZMA2 [16] | 2024 | C/C++ | LZ77, Dictionary-based Compression |
| XZ [7] | 2024 | C/C++ | LZ77, Dictionary-based Compression |
| GZip [43] | 2023 | C/C++ | LZ77, Dictionary-based Compression |
| PPMD [19] | 2022 | C/C++ | Prediction by Partial Match |
| SnZip [44] | 2021 | C/C++ | LZ77, Dictionary-based compression |
| Spring [45] | 2019 | C/C++ | BSC [46] and LZMA2 Compression |
| NAF [47] | 2019 | C/C++ | ZSTD [48] Compression Standard |

overall compression and decompression throughput [4, 9]:

$$THP = \frac{Size_{source}}{CT + DT}, \quad (4)$$

and compression robustness performance [2, 50, 52]:

$$CRP = \frac{\sqrt{\frac{1}{N} \times \sum_{i=0}^{N-1}(CR_i - CR_u)^2}}{CR_u} \times 100\%. \quad (5)$$

Among Eq.(4-5), the $CT$, $DT$, $CR_i$, $CR_u$, and $N$ denote compression time, decompression time, the $CR$ of the $i$-th tested file, average compression ratio, and the total number of datasets. Smaller $CR$ and $CRP$ mean better compression and robustness. Bigger $THP$ means faster speed in the compression and decompression process. Besides, we also evaluated the computing resource consumption for LLCs, including CPU and GPU peak memory usage [2, 20, 21, 50].

5.1.4 **Parameters Setting.** The SPuM is trained on the DNACorpus [10], which is a multi-source genomics dataset. The datasets OrSa, DiCo, EnIn, ScPo, and DaRe are used to fine-tune the super-parameters. Based on experiments and experience, we set $s = k = 3$ and $t = 32$ as the default configuration. It is important to note that the (3,3)-mer is a configuration that balances compression ratio and throughput, while other combinations still have advantages in either compression ratio or throughput. For baselines, we adopted the default settings from their papers or codes. All learning-based compressors used the same batch size ($bs = 320$).

### 5.2 Compression Ratio

As shown in Table 4, PMKLC-S and PMKLC-M achieved average compression ratio improvements ($\frac{Baseline-Ours}{Baseline} \times 100\%$) ranging from 0.557% ~ 73.609% and 0.072% ~ 73.480%, respectively. The main reason for this phenomenon is the introduction of the automated multi-knowledge learning-based framework, which enriches the knowledge sources and enhances the compression ratios.

We also noticed that TRACE [20], AGDLC [33], and MSDLC [41] have advantages in GD compression. The improvements of PMKLC-S/M in terms of Improve-S/M were 1.108%, 1.407%, and 0.557%, as well as 0.625%, 0.926%, and 0.072%, respectively. However, compared to those methods, PMKLC shown stronger advantages in throughput, CPU/GPU memory usage, and robustness performance. Taking MSDLC as an example, as shown in Table 5 and Table 6, the throughput of PMKLC-M is 2.034× faster that MSDLC, with an 83.967% savings in CPU memory (MSDLC consumes 19.335 GB, while PMKLC-M only requires 3.100 GB), an 8.269% reduction in GPU memory, and a 0.827% improvement in robustness.

Besides, we also presented a set of results showing the impact of model size on the CR, using DNA-BiLSTM [11] as an example, the model size of DNA-BiLSTM was included in the final compressed file, whereas DNA-BiLSTM* was not included. As DNA-BiLSTM shown, the CR was sensitive to the size of to-be-compressed datasets, with a CR of 12.523 bits/base for PlFa and 2.015 bits/base for TaGu. In contrast, PMKLC series demonstrates a more robust performance concerning dataset size. Table 6 provides quantitative results on compression ratio robustness, which we will analyze in the following subsections.

### 5.3 Throughput

Table 5 shows the overall compression and decompression throughput of PMKLC and Python-based learning-based lossless compressors. For PMKLC-S, the average throughput improvement Improv-S ($\frac{Ours-Baseline}{Baseline} \times 100\%$) reached 4.619% (MSDLC) ~ 303.578% (Lstm-compress). When using data block partitioning and SMP strategies for multi-GPU computing, the average throughput improvement (Improve-M) of PMKLC-M reached 203.569% ~ 1071.043%. Overall, both PMKLC-S and PMKLC-M were best in terms of compression and decompression throughput, indicating their great potential for real-time data compression. The throughput advantage of proposed PMKLC comes from three aspects: 1) GPU-accelerated $(s,k)$-mer encoding reduces data size and the GPU computing cores share the data processing; 2) automated model selection reduces the computational load of model inference; 3) Data block partitioning and Step-wise Model Passing strategies further accelerate the compression and decompression process.

Table 4: Compression Ratio ↓ (bits/base) of PMKLC and baselines. Improve-S/M represent the increase in average CR calculated based on our PMKLC-S/M. DNA-BiLSTM* indicates that the static model size is not included in the final compressed file

| Methods/Datasets | PlFa | WaMe | DrMe | GaGa | SnSt | MoGu | ArTh | AcSc | TaGu | Average | Improve-S (%) | Improve-M (%) |
|---|---|---|---|---|---|---|---|---|---|---|---|---|
| Spring | 1.859 | 1.988 | 1.939 | 1.904 | 1.886 | 1.680 | 1.934 | 1.877 | 1.883 | 1.883 | +2.035 | +1.558 |
| NAF | 1.874 | 1.988 | 1.990 | 1.959 | 1.967 | 1.799 | 1.929 | 1.905 | 1.943 | 1.928 | +4.316 | +3.849 |
| LZMA2 | 1.866 | 2.070 | 1.992 | 1.947 | 1.974 | 1.693 | 1.904 | 1.878 | 1.914 | 1.915 | +3.672 | +3.202 |
| XZ | 1.868 | 2.074 | 1.993 | 1.948 | 1.978 | 1.694 | 1.907 | 1.879 | 1.916 | 1.917 | +3.778 | +3.309 |
| PPMD | 1.906 | 2.109 | 2.054 | 2.026 | 1.995 | 1.780 | 2.008 | 1.972 | 1.998 | 1.983 | +6.964 | +6.511 |
| PBzip2 | 2.093 | 2.168 | 2.159 | 2.137 | 2.102 | 1.927 | 2.138 | 2.083 | 2.117 | 2.103 | +12.254 | +11.826 |
| Gzip | 2.120 | 2.250 | 2.220 | 2.187 | 2.195 | 2.030 | 2.185 | 2.140 | 2.163 | 2.166 | +14.802 | +14.387 |
| SnZip | 3.640 | 3.742 | 3.694 | 3.698 | 3.711 | 3.526 | 3.695 | 3.638 | 3.663 | 3.667 | +49.692 | +49.447 |
| DNA-BiLSTM | 12.523 | 12.428 | 4.904 | 2.559 | 17.235 | 2.421 | 6.780 | 2.053 | 2.015 | 6.991 | +73.609 | +73.480 |
| DNA-BiLSTM* | 1.856 | 1.946 | 1.926 | 1.914 | 1.908 | 1.861 | 1.913 | 1.919 | 1.924 | 1.907 | +3.274 | +2.802 |
| TRACE | 1.833 | 1.959 | 1.912 | 1.886 | 1.873 | 1.657 | 1.905 | 1.895 | 1.871 | 1.866 | +1.108 | +0.625 |
| DeepZip | 1.901 | 2.034 | 1.916 | 1.862 | 1.982 | 1.650 | 1.935 | 1.868 | 1.851 | 1.889 | +2.318 | +1.841 |
| AGDLC | 1.832 | 1.951 | 1.925 | 1.891 | 1.915 | 1.662 | 1.901 | 1.899 | 1.866 | 1.871 | +1.407 | +0.926 |
| MSDLC | 1.821 | 1.953 | 1.903 | 1.871 | 1.885 | 1.651 | 1.898 | 1.867 | 1.849 | 1.855 | +0.557 | +0.072 |
| Lstm-compress | 7.284 | 7.155 | 6.075 | 7.244 | 2.147 | 7.836 | 6.311 | 7.392 | 8.181 | 6.625 | +72.151 | +72.015 |
| JARVIS3 | 1.896 | 2.015 | 1.952 | 1.946 | 1.951 | 1.940 | 1.855 | 1.873 | 1.940 | 1.930 | +4.393 | +3.927 |
| **PMKLC-S (Ours)** | 1.812 | 1.943 | 1.900 | 1.850 | 1.851 | 1.651 | 1.892 | 1.866 | 1.844 | 1.845 | — | -0.464 |
| **PMKLC-M (Ours)** | 1.827 | 1.953 | 1.907 | 1.858 | 1.877 | 1.652 | 1.897 | 1.867 | 1.844 | 1.854 | +0.462 | — |

Table 5: Throughput ↑ (KB/S) of PMKLC and learning-based baselines. Improve-S/M is calculated based on ours PMKLC-S/M

| Methods/Datasets | PlFa | WaMe | DrMe | GaGa | SnSt | MoGu | ArTh | AcSc | TaGu | Average | Improve-S (%) | Improve-M (%) |
|---|---|---|---|---|---|---|---|---|---|---|---|---|
| DNA-BiLSTM | 12.588 | 12.595 | 12.426 | 11.119 | 12.668 | 10.915 | 12.611 | 6.837 | 5.427 | 10.799 | +94.349 | +463.934 |
| TRACE | 7.992 | 7.951 | 7.940 | 8.343 | 8.241 | 8.362 | 8.315 | 13.330 | 8.282 | 8.751 | +139.830 | +595.904 |
| DeepZip | 17.239 | 17.072 | 17.400 | 17.346 | 16.959 | 17.706 | 16.636 | 16.635 | 17.993 | 17.221 | +21.871 | +253.627 |
| AGDLC | 12.905 | 12.718 | 13.002 | 12.968 | 12.668 | 13.071 | 13.030 | 13.098 | 11.916 | 12.820 | +63.709 | +375.028 |
| MSDLC | 19.804 | 19.884 | 20.028 | 20.406 | 19.633 | 20.404 | 20.118 | 20.255 | 20.011 | 20.060 | +4.619 | +203.569 |
| Lstm-compress | 5.371 | 5.488 | 5.156 | 5.157 | 5.618 | 5.075 | 5.187 | 5.141 | 4.610 | 5.200 | +303.578 | +1071.043 |
| **PMKLC-S (Ours)** | 21.610 | 22.098 | 22.510 | 22.253 | 21.421 | 22.562 | 22.106 | 16.888 | 17.433 | 20.987 | — | +190.168 |
| **PMKLC-M (Ours)** | 60.941 | 60.699 | 71.416 | 75.581 | 58.070 | 75.457 | 69.664 | 38.156 | 38.093 | 60.897 | -65.537 | — |

Table 6: Compression Robustness Performance ↓ (%) and Average CPU/GPU Memory Usage ↓ (GB) of our PMKLC and learning-based baselines. The Improve-S/M and Saving-S/M are calculated based on ours PMKLC-S/M

| Metrics/Methods | DNA-BiLSTM | TRACE | DeepZip | AGDLC | MSDLC | **PMKLC-S** | **PMKLC-M** |
|---|---|---|---|---|---|---|---|
| **Compression Robustness Performance (%)** | 81.418 | 4.578 | 5.690 | 4.572 | 5.584 | 4.455 | 4.546 |
| Improve-S (%) | +94.528 | +2.687 | +21.705 | +2.559 | +2.814 | — | +2.002 |
| Improve-M (%) | +94.416 | +0.699 | +20.105 | +0.569 | +0.827 | -2.043 | — |
| **The Average CPU-Memory Usage (GB)** | 9.306 | 4.525 | 9.126 | 6.527 | 19.335 | 3.620 | 3.100 |
| Saving-S (%) | +61.100 | +20.000 | +60.333 | +44.538 | +81.277 | — | -16.774 |
| Saving-M (%) | +66.688 | +31.492 | +66.031 | +52.505 | +83.967 | +14.365 | — |
| **The Average GPU-Memory Usage (GB)** | 1.957 | 0.506 | 1.481 | 0.982 | 0.907 | 0.807 | 0.832 |
| Saving-S (%) | +58.763 | -59.486 | +45.510 | +17.821 | +11.025 | — | +3.005 |
| Saving-M (%) | +57.486 | -64.427 | +43.822 | +15.275 | +8.269 | -3.098 | — |

## 5.4 Compression Robustness

As shown in Table 6, PMKLC-S/M achieved the overall highest robustness performance. Compared to DNA-BiLSTM [11], TRACE [20], DeepZip [23], AGDLC [33], and MSDLC [41], the improvements ($\frac{Baseline-Ours}{Baseline} \times 100\%$) were 94.528%, 2.687%, 27.705%, 2.559%, and 2.814%, respectively. Compared to the baselines, the improvements of PMKLC-M were 94.416%, 0.699%, 20.105%, 0.569%, and 0.827%, respectively. This indicates that PMKLC demonstrates overall more robust performance when facing datasets with different probability distributions and scales, and it is minimally affected by perturbations in data probability distribution. We attribute this to two reasons: 1) The automated model selection component avoids static pre-training on small-scale datasets; 2) The multi-knowledge learning-based framework fully models the to-be-compressed datasets.

## 5.5 CPU & GPU Memory Usage

As shown in Table 6, for average CPU-memory usage, compared to baselines, the PMKLC-S/M saved ($\frac{Baseline-Ours}{Baseline} \times 100\%$) 20.000% (TRACE) ~ 81.277% (MSDLC) and 31.492% ~ 83.967%, respectively.

Table 7: The results of ablation study. C/GPM and MGPU denote C/GPU Peak Memory and Multiple GPU computing

|   | Ablation Module | | | | | | | OrSa (43,262,523 Bytes) | | | | DiCo (554,197,554 Bytes) | | | |
|---|---|---|---|---|---|---|---|---|---|---|---|---|---|---|---|
|   | SPuM | SPrM | DM | GskE | MS | SMP | MGPU | CR (bits/base) | THP (KB/S) | CPM (GB) | GPM (GB) | CR (bits/base) | THP (KB/S) | CPM (GB) | GPM (GB) |
| A | ✗ | ✗ | ✓ | ✗ | ✗ | ✗ | ✗ | 1.903 | 8.162 | 1.589 | 0.503 | 1.897 | 8.170 | 9.206 | 0.503 |
| B | ✗ | ✓ | ✓ | ✗ | ✗ | ✗ | ✗ | 1.993 | 5.921 | 3.606 | 3.220 | 1.884 | 5.934 | 34.109 | 3.220 |
| C | ✓ | ✓ | ✓ | ✗ | ✗ | ✗ | ✗ | 1.993 | 5.751 | 3.626 | 3.220 | 1.883 | 5.726 | 34.143 | 3.220 |
| D | ✓ | ✓ | ✓ | ✓ | ✗ | ✗ | ✗ | 2.022 | 16.967 | 1.873 | 3.222 | 1.882 | 16.896 | 12.059 | 3.222 |
| S | ✓ | ✓ | ✓ | ✓ | ✓ | ✗ | ✗ | 1.893 | 22.628 | 1.210 | 0.505 | 1.881 | 16.918 | 12.044 | 3.222 |
| E | ✓ | ✓ | ✓ | ✓ | ✓ | ✗ | ✓ | 1.901 | 82.506 | 1.047 | 0.505 | 1.885 | 40.337 | 12.047 | 3.222 |
| M | ✓ | ✓ | ✓ | ✓ | ✓ | ✓ | ✓ | 1.900 | 72.087 | 1.049 | 0.505 | 1.884 | 37.889 | 12.067 | 3.222 |

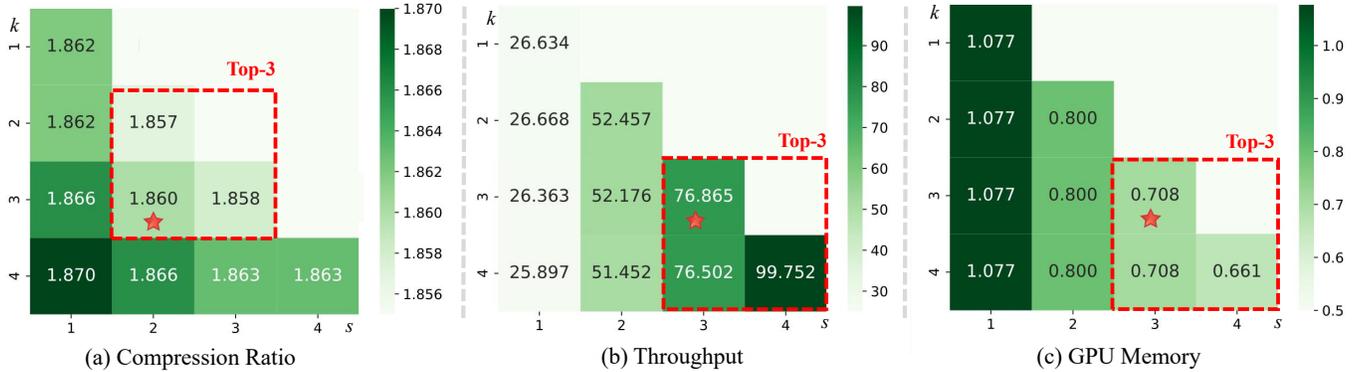

(a) Compression Ratio    (b) Throughput    (c) GPU Memory

Figure 6: The impact of different ($s,k$)-mer encoder on the GaGa dataset. "★" denotes the default configuration

Table 8: The impact of ($s,k$)-mer in time and memory on DiCo dataset. CPM and DPM denote CPU and GPU Peak Memory

|  | CPU-based ($s,k$)-mer | | GPU-based ($s,k$)-mer | | |
|---|---|---|---|---|---|
| ($s,k$)-mer | Time (S) | CPM (GB) | Time (S) | CPM (GB) | GPM (GB) |
| (1,1)-mer | 214.098 | 8.292 | 5.057 | 3.185 | 2.968 |
| (1,2)-mer | 223.001 | 8.292 | 5.046 | 3.185 | 2.968 |
| (2,2)-mer | 119.088 | 4.164 | 4.031 | 2.153 | 1.935 |
| (1,3)-mer | 221.053 | 8.292 | 5.046 | 3.185 | 2.968 |
| (2,3)-mer | 111.055 | 4.164 | 4.034 | 2.153 | 1.935 |
| (3,3)-mer | 75.039 | 2.787 | 3.095 | 1.809 | 1.591 |
| (1,4)-mer | 226.084 | 8.293 | 5.053 | 3.185 | 2.968 |
| (2,4)-mer | 123.021 | 4.164 | 4.039 | 3.185 | 1.935 |
| (3,4)-mer | 82.038 | 2.787 | 4.057 | 1.808 | 1.591 |
| (4,4)-mer | 63.040 | 2.100 | 3.077 | 1.637 | 1.419 |

For average GPU memory usage, PMKLC-S/M was only inferior to TRACE but outperformed DNA-BiLSTM, DeepZip, AGDLC, and MS-DLC. It saved GPU memory in the range of 11.025% ~ 58.763% and 8.269% ~ 57.468%, respectively. The memory advantage of PMKLC indicates its great potential on resource-constrained devices. This advantage is achieved through the combination of the automated model selector and the GPU-accelerated ($s,k$)-mer encoder.

### 5.6 Ablation Study

As shown in Table 7, we observe some interesting phenomena among several modules: 1) When the dataset was large (DiCo), multi-knowledge learning (MODE-A/B/C) enhanced the compression ratios but resulted in worse throughput and memory usage. 2) The introduction of the ($s,k$)-mer encoder optimized throughput and CPM (MODE-D) but also led to unstable variations in CR. 3) The introduction of the MS ensured that MODE-S achieved an overall balance of compression ratio, memory, and throughput, mainly because it reduced the model inference and considered the impact of data size. 4) Mode-M achieved higher throughput at the cost of a slight loss in compression ratio; meanwhile, the introduction of SMP alleviated the cold-start issue (MODE-E), providing further balance between compression ratio and throughput.

### 5.7 The Impact of ($s,k$)-mer Encoder

Table 8 shows the runtime and memory cost of CPU/GPU-based ($s,k$)-mer on the DiCo dataset. The GPU-accelerated ($s,k$)-mer achieved a maximum speedup of 44.743× while requiring the overall lowest memory consumption. Besides, as shown in Figure 6, the combinations of (2,2), (3,2), (3,3) achieved the overall Top-3 performance on CR, THP, and GPU memory usage, considering the overall balance, the (3,3)-mer was used as the default configure of PMKLC.

### 5.8 The Impact of Batch Size $bs$

As shown in Fig. 7(a-b), the CR of PMKLC worsen as the batch size ($bs$) increases, while the THP gradually increases within a certain range. Considering the balance between compression ratio and throughput, we set $bs = 320$ as the default setting.

### 5.9 The Impact of Context Length $t$

As shown in Fig. 7(c-d), a larger context length ($t$) means a stronger compression ratio, but at the cost of reduced compression and decompression throughput. In this study, we set $t = 32$ to balance compression ratio and throughput; however, other values of $t$ are still valid and can be chosen according to the user's needs.

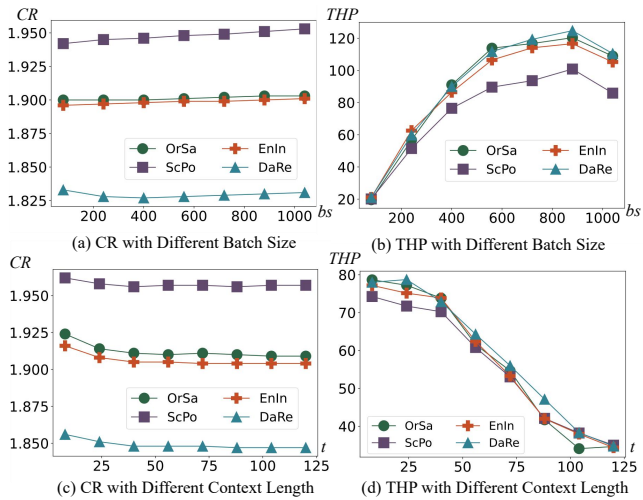

**Figure 7: The impact of different batch size ($bs$) and context length ($t$) on datasets OrSa, EnIn, ScPo, and DaRe**

## 6 Conclusion

In this study, we propose PMKLC, a parallel multi-knowledge learning-based compressor for balancing compression ratio, robustness, throughput, and memory usage. We develop two compression modes, PMKLC-S and PMKLC-M, and validate their effectiveness against 14 baselines. We evaluate the performance of the two models across multiple assessment metrics, including compression ratio, throughput, robustness, and memory consumption. The experiments show that the compression ratio improvement of PMKLC-S/M up to 73.609% and 73.480%, while the throughput improvement up to 303.578% and 1071.043%, respectively. Additionally, PMKLC-S/M demonstrated optimal compression robustness and competitive memory advantages. The compression robustness performance improvements for PMKLC-S and PMKLC-M are 94.528% and 94.416%, respectively. The average maximum CPU memory savings are 81.277% and 83.967%, respectively, while the average maximum GPU memory savings are 58.763% and 57.486%, respectively. In the future, our purpose is to design parallel compression framework aimed at large-scale biological data, including lossy medical images and quality score data.

## 7 Acknowledgments

This work was partly supported by the National Natural Science Foundation of China under Grant (62272253, 62272252) and the China Scholarship Council (CSC) scholarship program.